\documentclass[mlmain]{jmlr}

\usepackage{longtable}% for long tables

\usepackage{booktabs}
\usepackage[load-configurations=version-1]{siunitx} % newer version

\usepackage{wrapfig}

\usepackage[skip=0pt]{caption}

\theorembodyfont{\upshape}
\theoremheaderfont{\scshape}
\theorempostheader{:}
\theoremsep{\newline}

\jmlrvolume{}
\firstpageno{1}
\editors{Simone Azeglio, Christian Shewmake, Bahareh Tolooshams, Sophia Sanborn, Chase van de Geijin, Nina Miolane}

\jmlryear{2024}
\jmlrworkshop{Symmetry and Geometry in Neural Representations}

\title[Dynamical symmetries in the fluctuation-driven regime]{Dynamical symmetries in the fluctuation-driven regime: an application of Noether's theorem to noisy dynamical systems}

\author{\Name{John J. Vastola} \Email{john\_vastola@hms.harvard.edu \\} \addr Department of Neurobiology, Harvard Medical School, Boston, MA, USA}

\begin{document}

\maketitle

\begin{abstract}
Noether's theorem provides a powerful link between continuous symmetries and conserved quantities for systems governed by some variational principle. Perhaps unfortunately, most dynamical systems of interest in neuroscience and artificial intelligence cannot be described by any such principle. On the other hand, nonequilibrium physics provides a variational principle that describes how fairly generic noisy dynamical systems are most likely to transition between two states; in this work, we exploit this principle to apply Noether's theorem, and hence learn about how the continuous symmetries of dynamical systems constrain their most likely trajectories. We identify analogues of the conservation of energy, momentum, and angular momentum, and briefly discuss examples of each in the context of models of decision-making, recurrent neural networks, and diffusion generative models. 
\end{abstract}

\section{Introduction}
\label{sec:intro}

In physics, Noether's theorem provides a fundamental link between the symmetries of physical systems on the one hand, and conserved quantities like energy and momentum on the other hand \citep{Noether1918,Kosmann2011-book,Neuenschwander2017-wonderful}. In its modern form, it uniquely associates (equivalence classes of) independent continuous symmetries, which can be formalized in terms of Lie groups and algebras, with (equivalence classes of) independent conserved quantities \citep{MartinezAlonso1979,Olver1986-collection,Olver1993applications,Brown2020-history}. Importantly, it allows us to make nontrivial statements about the dynamics of systems which may be otherwise extremely difficult to analyze. 

The utility of Noether's theorem has inspired a variety of extensions and applications, including in the context of classical mechanics \citep{Sarlet1981-CMgeneralizations}, non-variational partial differential equations \citep{Anco2017}, statistical mechanics \citep{Hermann2021}, and Markov models \citep{Baez2013-markov}. But it is worth emphasizing that the physical systems considered in typical applications of Noether's theorem---whose dynamics can be characterized as extremizing a certain functional---are extremely special. Most dynamical systems, including those most relevant to neuroscience and artificial intelligence (AI), \textit{cannot} be described in terms of some nontrivial variational principle. Moreover, many systems of interest exhibit features that suggest they are not even `almost' describable by some variational principle. For example, most dynamical systems lack a notion of `inertia': their dynamics are completely determined by their current state, rather their current state and some notion of state velocity. Additionally, most dynamical systems are not `gradient-like', since the function governing how states evolve in time is not the gradient of some scalar potential. Either feature can yield major qualitative differences in dynamics: for example, `normal' versus `non-normal' dynamics \citep{Hennequin2012,Kerg2019-nonnormal} maps onto the gradient-like versus non-gradient-like distinction.

Nonetheless, one still hopes that symmetries can be exploited to say \textit{something} about the dynamics of many systems of interest, including neural networks. Of special note is the work of \citet{Tanaka2021-noether-learning}, which says that neural network learning dynamics can be associated with a certain Lagrangian when gradient descent is performed in discrete steps, and hence that a version of Noether's theorem holds. But what can be said about more general dynamics not necessarily tied to learning? Consider, for example, the firing rate dynamics of some trained recurrent neural network (RNN). Given symmetries, what can we say about the network's trajectory through state space? 

Perhaps surprisingly, there is a sense in which Noether's theorem applies to extremely general noisy dynamical systems, including noisy RNNs and diffusion models. In this work, our goal is to explore this application and interpret the associated conserved quantities. The idea is that the transient and fluctuation-driven behavior (e.g., jumps between neighboring attractor basins) of such systems is well-described by a certain variational principle, which allows us to apply Noether's theorem to learn about how state transitions are most likely to occur. Noether's theorem tells us that even though noise plays a crucial role in many of these transitions, their form is not arbitrary, but in fact constrained by a system's symmetries.  

The paper is organized as follows. First, we review Noether's theorem and some of its well-known consequences. Next, we (re-)derive a variational principle that allows us to apply Noether's theorem to noisy dynamical systems. Finally, we discuss various consequences of this application of Noether's theorem, including analogues of energy and momentum conservation. We consider a few example systems relevant to neuroscience and AI, including drift-diffusion models of decision-making, noisy RNNs, and diffusion generative models.

%%%%%%%%%%%%%%%%%%%%%%%%%%%%%%

\section{Noether's theorem, symmetries, and conserved quantities}

\label{sec:prelim}

What is colloquially referred to as ``Noether's theorem'' was first presented in a paper by Emmy Noether in 1918 \citep{Noether1918}. Two subtleties are worth mentioning: first, Noether's paper actually contains two theorems, and ``Noether's theorem'' without additional qualifications usually refers to the first; second, although the theorem is usually invoked to show that continuous symmetries imply conserved quantities, the converse direction (conserved quantity $\implies$ symmetry) actually also holds when one is careful about how one defines ``symmetry'' and ``conserved quantity''. See \citet{Brown2020-history} and \citet{Olver2018emmy} for helpful discussions of the history and precise formulation of the converse direction.

For our purposes, the simple version of Noether's (first) theorem presented in physics textbooks on mechanics and field theory will suffice. The standard story is usually told in terms of actions and Lagrangians; while an alternative formulation which exploits a Hamiltonian rather than Lagrangian viewpoint is also available, and arguably simplifies the picture in a number of ways \citep{Baez2020-bottom}, we will not consider it here. Since our applications involve trajectories parameterized only by one variable, time, we present the `particle' rather than the `field' version of Noether's theorem.

Noether's theorem applies to systems described by a variational principle. Let $\vec{x}(t) := ( x_1(t), ..., x_N(t) )^T \in \mathbb{R}^N$ denote the state of one such system at time $t \in [t_0, t_f]$. We assume that the system's trajectory $\vec{x}(t)$ extremizes the `action' functional
\begin{equation}
\mathcal{S}[ \vec{x} ] := \int_{t_0}^{t_f} L( \vec{x}(t), \dot{\vec{x}}(t), t) \ dt 
\end{equation}
where the function $L$ is the corresponding Lagrangian. By a standard calculus of variations result, this assumption implies the system's trajectory satisfies the Euler-Lagrange equations
\begin{equation}
\frac{\partial L}{\partial \vec{x}} = \frac{d}{dt}\left( \frac{\partial L}{\partial \dot{\vec{x}}} \right) \hspace{0.3in}  \text{or equivalently} \hspace{0.3in} \frac{\partial L}{\partial x_i} = \frac{d}{dt}\left( \frac{\partial L}{\partial \dot{x}_i} \right) \ \text{ for } i = 1, ..., N \ .
\end{equation}
A continuous symmetry of such a system is a continuous transformation that leaves the action $\mathcal{S}$ invariant (regardless of whether the equations of motion are satisfied). The more general notion of a continuous \textit{quasi}-symmetry refers to a continuous transformation that leaves the action \textit{almost} invariant---i.e., up to a surface term. Noether's theorem is about the consequences of quasi-symmetry given that the equations of motion are satisfied. 

\begin{theorem}[Noether's theorem]\label{thm:noether}
Let $\delta t \geq 0$ and $\delta \vec{x} \in \mathbb{R}^N$, and consider a transformation that takes $t \to t' := t + \epsilon \ \delta t$ and $\vec{x} \to \vec{x}' := \vec{x} + \epsilon \ \delta \vec{x}$. If the Lagrangian $L$ satisfies
\begin{equation}
L( \vec{x}', \dot{\vec{x}}', t' ) = L( \vec{x}, \dot{\vec{x}}, t ) + \frac{d}{dt} K(\vec{x}, \dot{\vec{x}}, t) + \mathcal{O}(\epsilon^2) 
\end{equation}
for all $\epsilon > 0$ sufficiently small, where $K$ is some function, then the quantity
\begin{equation}
J := \sum_i \frac{\partial L}{\partial \dot{x}_i} \delta x_i - K
\end{equation}
is conserved in the sense that $dJ/dt = 0$ when the equations of motion are satisfied.
\end{theorem}
The proof of this version of Noether's theorem is standard, so we do not include it here. Well-known consequences of Noether's theorem include the conservation of energy, which follows from (quasi-)symmetry with respect to time translations; the conservation of momentum, which follows from symmetry with respect to translations along one or more directions of state space; and the conservation of angular momentum, which follows from symmetry with respect to state space rotations within one or more planes.

Importantly, although the aforementioned quantities have specific forms for typical physical systems---e.g., energy is kinetic plus potential energy, and momentum is mass times velocity---one can derive \textit{analogues} of the more familiar energy, momentum, and so on as long as the associated symmetries still hold for a non-physical system that can be characterized by a Lagrangian. We exploit this fact in what follows.

\section{Applying Noether's theorem to stochastic dynamical systems}
\label{sec:derivation}

Let us now consider a system defined by a set of stochastic differential equations (SDEs)
\begin{equation} \label{eq:SDE_general}
\dot{\vec{x}} = \vec{f}(\vec{x}, t) + \vec{G}(\vec{x}, t) \ \vec{\eta}_t
\end{equation}
where $\vec{x}(t) := ( x_1(t), ..., x_N(t) )^T \in \mathbb{R}^N$ denotes the system's state at time $t \in [t_0, t_f]$, $\vec{f}: \mathbb{R}^N \times [t_0, t_f] \to \mathbb{R}^N$ determines the drift term, $\vec{G}: \mathbb{R}^N \times [t_0, t_f] \to \mathbb{R}^{N \times M}$ determines the noise amplitude, and $\vec{\eta}_t$ is an $M$-dimensional vector of independent Gaussian white noise terms. For technical reasons, we assume that the $N \times N$ diffusion tensor $\vec{D}(\vec{x}, t) := \frac{1}{2} \vec{G}(\vec{x}, t) \vec{G}(\vec{x}, t)^T$ is positive definite for all $\vec{x}$ and $t$, and hence always invertible.

To apply Noether's theorem to systems described by Eq. \ref{eq:SDE_general}, we must identify a relevant variational principle. But from the form of Eq. \ref{eq:SDE_general} alone, it is not clear why we should expect one to exist, or what form it should take if it does. A path forward becomes clearer if one exploits the rich analogy between quantum mechanics and stochastic dynamics. 

In one view, quantum mechanical systems are characterized by the time evolution of their wave function $\psi(\vec{x}, t)$, which encodes the probability of each possible state occupancy observation. The probability \textit{amplitude} associated with a transition from state $\vec{x}_0$ at time $t_0$ to a state $\vec{x}$ at time $t_f$ can be written as \citep{feynman2010quantum,kleinert2006path}
\begin{equation}
K( \vec{x}, t_f; \vec{x}_0, t_0) = \int \mathcal{D}[\vec{x}] \exp\left\{ \frac{i}{\hbar} \mathcal{S}[\vec{x}] \right\}
\end{equation}
where the integral is over all possible paths from $\vec{x}_0$ to $\vec{x}$, and where $\mathcal{S}$ is the system's associated action functional. What makes this particular description of a quantum system's behavior interesting is that it makes the connection between quantum and classical mechanics more explicit: each possible path from $\vec{x}_0$ to $\vec{x}$ contributes to the above path integral, but the path that extremizes $\mathcal{S}$---i.e., the classical path---contributes the most, and dominates in the $\hbar \to 0$ limit, or equivalently when quantum effects are taken to be negligible. Said differently, the above path integral description provides us with a variational principle that describes what path the system is most likely to take from $\vec{x}_0$ to $\vec{x}$.

A path integral description of stochastic dynamics also exists, and says the probability (rather than probability amplitude) associated with an $\vec{x}_0$ to $\vec{x}$ transition can be written
\begin{equation} \label{eq:L_general}
\begin{split}
p( \vec{x}, t_f; \vec{x}_0, t_0) = \int \mathcal{D}[\vec{x}] \exp\left\{ - \mathcal{S}[\vec{x}] \right\} \hspace{1in} \mathcal{S}[\vec{x}] := \int_{t_0}^{t_f} L(\vec{x}, \dot{\vec{x}}, t) \ dt  \\
L(\vec{x}, \dot{\vec{x}}, t) := \frac{1}{4} \ [ \dot{\vec{x}} - \vec{f}(\vec{x}, t) ]^T \vec{D}^{-1}(\vec{x}, t) [ \dot{\vec{x}} - \vec{f}(\vec{x}, t) ] \ .
\end{split}
\end{equation}
This construction is usually called the Onsager-Machlup path integral \citep{OM1953,Cugliandolo_2017}, and is essentially equivalent to the Martin-Siggia-Rose \citep{MSR1973} path integral. The Lagrangian $L$ here is a quadratic form which can be interpreted as follows: the system \textit{usually} follows its drift term, but can fluctuate in other directions, with the strength and direction of those fluctuations being controlled by the diffusion tensor. In directions with higher noise, fluctuations incur a smaller action cost, and hence they are more likely. Unlike in the case of quantum and classical mechanics, the most likely path is the one which \textit{minimizes} $\mathcal{S}$. This path dominates the path integral in the small noise limit---yielding a stochastic dynamics analogue to classical mechanics---as one can formally show by exploiting Laplace's method. 

Now that we have a variational principle associated with Eq. \ref{eq:SDE_general}, we can apply Noether's theorem. In the next section, we introduce analogues of familiar conserved quantities.

\section{Energy, momentum, and angular momentum conservation}

Applying Noether's theorem to our Lagrangian (Eq. \ref{eq:L_general}) tells us how a system's continuous symmetries and quasi-symmetries constrain the most likely path it takes between any two states, including paths that require noise, like transitions between attractor basins. Since this Lagrangian can exhibit some of the same (quasi-)symmetries as physical systems, one can derive analogues of energy, momentum, and angular momentum conservation. Given that we started with a fairly arbitrary stochastic dynamical system (e.g., $\vec{f}$ is not necessarily the gradient of any potential function), this may be somewhat surprising.

A caveat, however: one should not get too greedy. While one might hope that applying Noether's theorem here produces conserved quantities that remain interesting as one takes the noise in the system to zero, since many dynamical systems of interest are deterministic, we will see that this does not happen. All of the conservation laws we obtain reduce to trivialities in the zero noise limit, which might be expected in light of the fact that we are not making any assumptions about $\vec{f}$. One cannot get something from nothing, after all.

\subsection{Energy conservation}

Assume our Lagrangian does not depend explicitly on time, which is equivalent to demanding that neither $\vec{f}$ nor $\vec{D}$ depends explicitly on time. Then an infinitesimal time translation $t' := t + \epsilon$ produces a state perturbation $\vec{x}(t') = \vec{x}(t) + \dot{\vec{x}}(t) \ \epsilon$ and changes $L$ as
\begin{equation}
L \to L + \epsilon \left[ \frac{\partial L}{\partial \vec{x}} \cdot \dot{\vec{x}} + \frac{\partial L}{\partial \dot{\vec{x}}} \cdot \ddot{\vec{x}}   \right] = L + \epsilon \left[ \frac{dL}{dt} - \frac{\partial L}{\partial t}  \right] = L + \epsilon \frac{dL}{dt}
\end{equation}
where we used the fact that $L$ does not depend explicitly on time in the last step. By Noether's theorem, the quantity
\begin{equation}
E :=  \frac{\partial L}{\partial \dot{\vec{x}}} \cdot \dot{\vec{x}} - L = \frac{1}{4}  \dot{\vec{x}}^T \vec{D}^{-1} \dot{\vec{x}} - \frac{1}{4} \vec{f}^T \vec{D}^{-1} \vec{f}
\end{equation}
is conserved, i.e., $dE/dt = 0$ when the equations of motion are satisfied. Considering special cases helps us make sense of this quantity. Suppose $\vec{D}$ is a diagonal matrix and independent of $\vec{x}$. If $\vec{f} = - \vec{D} \frac{\partial V}{\partial \vec{x}}$ for some scalar potential $V(\vec{x})$, our original $L$ is equivalent to
\begin{equation}
L' = \frac{1}{4} \ \dot{\vec{x}} \vec{D}^{-1} \dot{\vec{x}} + \frac{1}{4} \ \vec{f} \vec{D}^{-1} \vec{f} \ ,
\end{equation}
since the cross term $- \frac{1}{2} \dot{\vec{x}} \frac{\partial V}{\partial \vec{x}}$ is equal to a total time derivative, and hence does not contribute to the equations of motion. This Lagrangian `looks' like a kinetic energy term (the mass associated with $x_i$ is $\frac{1}{2 D_{ii}}$) and a potential term $W(\vec{x}) := \frac{1}{4} \vec{f}(\vec{x})^T \vec{D}^{-1} \vec{f}(\vec{x})$, except that the potential term has the `wrong' sign. Similarly, our expression for $E$ becomes kinetic \textit{minus} potential energy, which is opposite how $E$ looks for physical systems. Another informative special case assumes $\vec{D}$ is state-independent and isotropic, i.e., $\vec{D} = D \vec{I}_N$, so
\begin{equation}
E = \frac{\Vert \dot{\vec{x}} \Vert_2^2 - \Vert \vec{f}(\vec{x}) \Vert_2^2}{4 D} =  \sum_i \frac{\dot{x}_i^2 - f_i(\vec{x})^2}{4 D} \ .
\end{equation}
The above expression makes clear that the conservation of energy is a statement about how the system's fluctuation-driven trajectory differs from the deterministic trajectory determined only by its drift term $\vec{f}$. The squared norm of the `state velocity' $\dot{\vec{x}}$ deviates from that of the drift term in general, but \textit{by the same amount} throughout its most likely transition between two states. When the system follows its deterministic trajectory, $E = 0$.

Since $E$ is greater when the difference between the norms of $\dot{\vec{x}}$ and $\vec{f}(\vec{x})$ is large, we might expect the size of $E$ to be related to the time the system takes to transition between two states. (As a related point, note that $E$ has units of inverse time.) This turns out to be true, and for 1D systems one has a one-to-one relationship through the easy-to-derive 
\begin{equation}
t_* = \int_{x_0}^{x_f} \frac{dx}{\sqrt{f(x)^2 + 4 D(x) E}} \ .
\end{equation}
In other words, the transition time $t_*$ is strictly decreasing as a function of energy.

\subsection{Momentum conservation}

Assume our Lagrangian does not depend explicitly on $x_i$, i.e., that neither $\vec{f}$ nor $\vec{D}$ does. Then an infinitesimal translation $x_i' \to x_i + \epsilon$ along direction $i$ does not change $L$, and 
\begin{equation}
p_i := \frac{\partial L}{\partial \dot{x_i}} = \frac{1}{2} \sum_{j} D^{-1}_{ij} \left[ \dot{x}_j - f_j(\vec{x}, t)  \right]
\end{equation}
is conserved. More generally, momentum is defined as the $N$-dimensional vector
\begin{equation}
\vec{p} := \frac{\partial L}{\partial \dot{\vec{x}}} = \frac{1}{2} \vec{D}^{-1}(\vec{x}, t) \left[ \dot{\vec{x}} - \vec{f}(\vec{x}, t) \right] 
\end{equation}
and can be conserved in up to $N$ independent directions. While energy appears to measure the overall extent to which the deterministic dynamics are violated, different components of $\vec{p}$ quantify how badly this happens along each direction of state space. As in the case of energy, when the system follows its deterministic trajectory, $\dot{\vec{x}} = \vec{f}$, so $\vec{p} = \vec{0}$.

\subsection{Angular momentum conservation}

Assume our Lagrangian is invariant to rotations within the $x_i$-$x_j$ plane, and in particular to infinitesimal rotations of the form $x_i' = x_i - \epsilon x_j$, $x_j' = x_j + \epsilon x_i$. By Noether's theorem,
\begin{equation}
L_{ij} := \frac{\partial L}{\partial \dot{x}_i} \delta x_i + \frac{\partial L}{\partial \dot{x}_j} \delta x_j = x_i p_j - x_j p_i 
\end{equation}
is conserved. More generally, we can define an $N \times N$ angular momentum tensor 
\begin{equation}
\vec{L} := \vec{x} \vec{p}^T - \vec{p} \vec{x}^T = \frac{1}{2} \vec{x} \vec{D}^{-1} ( \dot{\vec{x}} - \vec{f} )^T - \frac{1}{2} ( \dot{\vec{x}} - \vec{f} ) \vec{D}^{-1} \vec{x}^T 
\end{equation}
which is antisymmetric and has up to $N (N-1)/2$ distinct entries (not coincidentally, the number of generators of the rotation group $\text{SO}(N)$), each of which may be conserved. The conservation of $L_{ij}$ implies that the extent to which dynamics deviate from the deterministic trajectory along the $x_i$ direction constrains the extent to which they can deviate along the $x_j$ direction. When the system follows its drift term, $p_i = p_j = 0$ and hence $L_{ij} = 0$.

\section{Case studies relevant to neuroscience and AI}

We now apply Noether's theorem to three example settings: (i) drift-diffusion models of decision-making,
 (ii) RNNs with point attractors, and (iii) diffusion generative models. We compute most likely transition paths directly via action minimization \citep{Strang2023}.\footnote{See \url{https://github.com/john-vastola/noether-neurreps24} for code.}

\begin{figure}[htbp]
 % Caption and label go in the first argument and the figure contents
 % go in the second argument
\floatconts
  {fig:decision}
  {\caption{Conservation laws relevant to simple decision-making and decision memory models. \textbf{a.} For a drift-diffusion model with fixed decision bounds at $\pm 1$, the most likely path from any evidence state to the opposite boundary is (approximately) a straight line (blue, `LAP'). Raw paths ($n = 981$) with same start state, end state, and transition time also shown. \textbf{b.} Transitions between two states of a single-attractor model (heatmap: steady state distribution) are direct when energy and angular momentum are high, and involve a diversion to the attractor state when they are low.  \textbf{c.} Transitions between two attractor basins in a model with three attractors (heatmap: steady state distribution) are more direct when energy is high, and involve visiting an intermediate attractor when energy is low.}}
  {\includegraphics[width=0.9\linewidth]{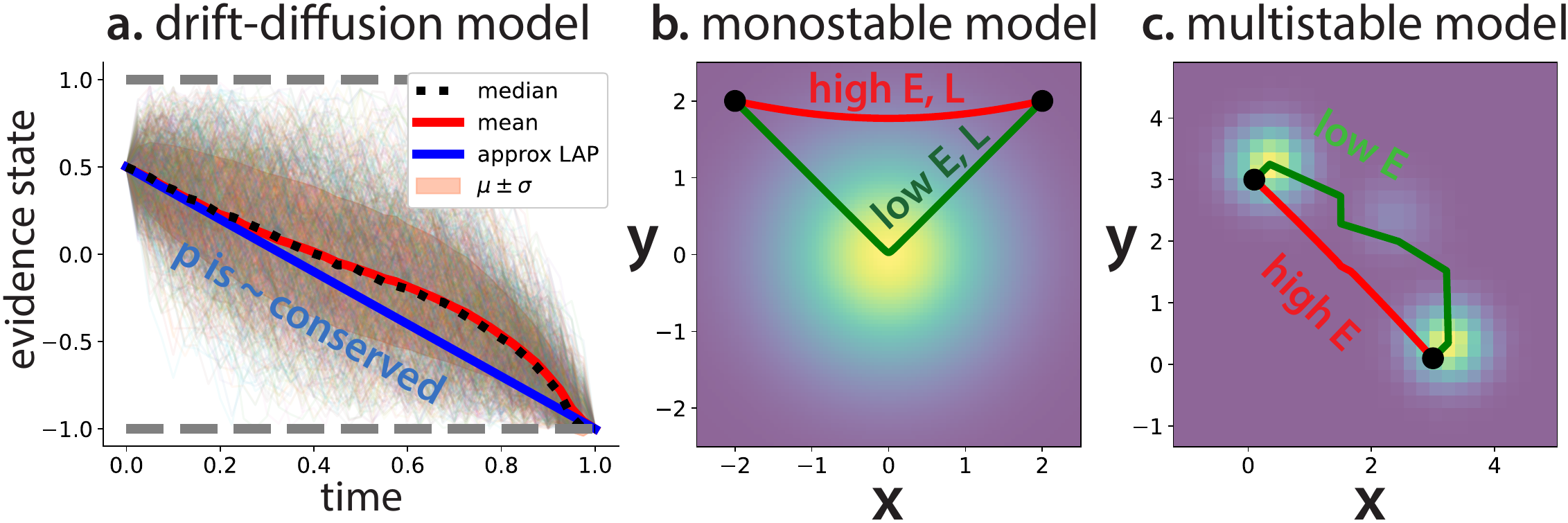}}
\end{figure}

\subsection{Drift-diffusion models of decision-making}

Drift-diffusion models \citep{Ratcliff2016} purport to describe the dynamics of decision-making in various multiple-alternative forced choice tasks (e.g., are dots on a screen moving mostly to the left or right?). They involve two components: Brownian-like particle dynamics 
\begin{equation}
\dot{x} = v + \sigma \ \eta_t  \hspace{1in} t \in [0, \infty)
\end{equation}
where $v = \text{const.}$ is the drift term (or `bias') and $\sigma$ is the noise strength, and boundary conditions that specify how the particle's state determines a decision. In the simplest case, which we will assume here, a decision happens when $x$ reaches either an upper or lower threshold, say, at $\pm 1$. More complex boundary shapes are often considered, e.g., time-dependent bounds that model decision urgency. But even simple drift-diffusion models explain available behavioral and neural data surprisingly well \citep{Gold2007}. 

The decision usually matches the sign of the drift term, but fluctuations, which can be viewed as representing random variations in the decision-maker's incoming evidence, can cause the particle to move in the opposite direction. One can ask: given that the system makes the `wrong' decision at some time $T \geq 0$, what path through state space did the system most likely take? Or equivalently: did the decision-maker most likely receive a sudden burst of contrary evidence, a steady stream of it, or something else?

For $f = v$ and $D = \sigma^2/2$, if we neglect decision-boundary-related effects, we have a Lagrangian that does not depend explicitly on $x$, so momentum is conserved. Then
\begin{equation}
p = \frac{\dot{x} - v}{\sigma^2} = \text{const.} \ \implies \ x(t) = x_0 + (v + \sigma^2 p) t = x_0 + \frac{(x_f - x_0)}{T} t \ ,
\end{equation}
i.e., that the system is most likely to travel from $x_0$ to $x_f$ in an amount of time $T$ with a constant velocity. In other words, the most likely way the decision-maker makes the wrong choice is by unluckily receiving a steady stream of contrary evidence. In practice, the presence of the decision boundaries modifies this conclusion, but only slightly (Fig. \ref{fig:decision}a).

\subsection{Changes of mind in an attractor network model of decision memory}

Animal decision-making is thought to involve at least three processes: (i) evidence accumulation, (ii) weighing evidence to come to a decision, and (iii) storing that decision for later use. Attractor network models, where each possible decision corresponds to a different attractor basin, provide a popular possible means by which neural circuits could achieve some combination of the above three aims, but especially the last \citep{Machens2005,Kopec2015,Piet2017-ncomp-dm}. 

Our conservation laws provide a way to think about movement both within and between basins, assuming no explicit time-dependence. Consider a 2D Ornstein-Uhlenbeck process as a model of within-attractor dynamics, and note both energy $E$ and angular momentum $L$ are conserved. State transitions tend to be more direct when $E$ and $L$ are high (i.e., when the transition is fast), and to visit the attractor's fixed point when they are low (Fig. \ref{fig:decision}b). 

Interestingly, multistable attractor network models predict that `changes of mind' are possible depending on the distance between and relative shallowness of different attractor basins \citep{Albantakis2011}. For example, if the available evidence is somewhat ambiguous during a two-alternative forced choice task, the basins may be relatively shallow and close together, and one might expect a relatively high rate of `hops' between them. 

Consider a modified version of the model from \citet{Piet2017-ncomp-dm} with
\begin{equation}
\begin{split}
\tau \dot{x} &= \mu_0 + \frac{A}{2} \left[ \tanh(\frac{x-c}{n}) + 1 \right] - \frac{I}{2} \left[ \tanh(\frac{y-c}{n}) + 1 \right] - x + \sqrt{2 \tau} \sigma \ \eta_x  \\
\tau \dot{x} &= \mu_0 + \frac{A}{2} \left[ \tanh(\frac{y-c}{n}) + 1 \right] - \frac{I}{2} \left[ \tanh(\frac{x-c}{n}) + 1 \right] - y + \sqrt{2 \tau} \sigma \ \eta_y  
\end{split}
\end{equation}
which is a multistable attractor network. The parameters $\mu_0$, $A$, $I$, and $\sigma$ control the size and number of attractor basins. We consider a parameter set with three stable states, and interpret one as an `undecided' state. Given a transition between the two choice-related basins, will this system stop at the `undecided' state, or not? The answer depends on the system's energy $E$. When $E$ is high, the system takes a direct path; when $E$ is low, the system must visit the intermediate state en route to the other choice basin (Fig. \ref{fig:decision}c).

\subsection{Data manifold symmetries and reverse diffusion dynamics}

Diffusion models sample from realistic and high-dimensional data distributions \citep{sohl-dickstein2015noneq,song2021scorebased,yang2022review}, and generalize well \citep{wang2024the,vastola2025generalization}. The core idea is to corrupt a data distribution via a forward stochastic process, and then learn to reverse the mapping in order to generate samples from noise: 
\begin{align}
\dot{\vec{x}} &= \sqrt{2 t} \ \vec{\eta}_t && t = 0 \rightarrow t = T &&& \text{forward process, } p_{data} \text{ to } p_{noise}   \\
\dot{\vec{x}} &= - t \ \vec{s}(\vec{x}, t) && t = T \rightarrow t = 0 &&& \text{reverse process A, } p_{noise} \text{ to } p_{data} \notag  \\
\dot{\vec{x}} &= - 2 t \ \vec{s}(\vec{x}, t) + \sqrt{2 t} \ \vec{\eta}_t  && t = T \rightarrow t = 0 &&& \text{ \ reverse process B, } p_{noise} \text{ to } p_{data}   \notag 
\end{align}
Here, $p(\vec{x} | t)$ is the noise-corrupted data distribution and $\vec{s}(\vec{x}, t) := \nabla_{\vec{x}} \log p(\vec{x} | t)$ is the score function. Two reverse processes are shown: the popular probability-flow ODE (PF-ODE), and the reverse SDE, which involves noise \citep{song2021scorebased}. In practice, either can outperform the other sample-quality-wise depending on the data set \citep{karras2022elucidating}. How do the reverse diffusion trajectories of the two approaches differ?  

We can exploit (rotation) symmetry to provide an answer in an interesting special case. Consider a Gaussian defined on a 2D ring of radius $R$, whose noise-corrupted version is
\begin{equation*}
p_{data}(\vec{x} | t) :=  \frac{1}{2 \pi (\sigma_0^2 + t^2)} \int_0^{2\pi} e^{ - \frac{(x - R \cos\theta)^2}{2 (\sigma_0^2 + t^2)} - \frac{(y - R \sin\theta)^2}{2 (\sigma_0^2 + t^2)} } \ \frac{d\theta}{2 \pi} = \frac{e^{ - \frac{(\Vert \vec{x} \Vert_2^2 + R^2)}{2 (\sigma_0^2 + t^2)} } }{2 \pi t^2} \ I_0\left( \frac{R}{\sigma_0^2 + t^2} \Vert \vec{x} \Vert_2 \right) 
\end{equation*}
where $I_0$ denotes the modified Bessel function of the first kind. One can use this expression to analytically compute the score function. One finds that PF-ODE trajectories are not curved, but move towards the ring along the line perpendicular to its nearest tangent. Meanwhile, the stochastic sampler can produce trajectories that end at various other locations. The differences between these trajectories can be conveniently characterized in terms of angular momentum $L$, which is conserved by the rotation symmetry of the data distribution, and hence of the score; reaching points near the deterministic trajectory's end is `easier', and hence requires less angular momentum, while reaching points on the other side of the ring requires greater angular momentum (Fig. 2).

\begin{wrapfigure}{r}{0.5\textwidth}
\vspace{-0.5in}
 % Caption and label go in the first argument and the figure contents
 % go in the second argument
%\floatconts
 % {fig:revdiff}
  \begin{center}
  \includegraphics[width=0.85\linewidth]{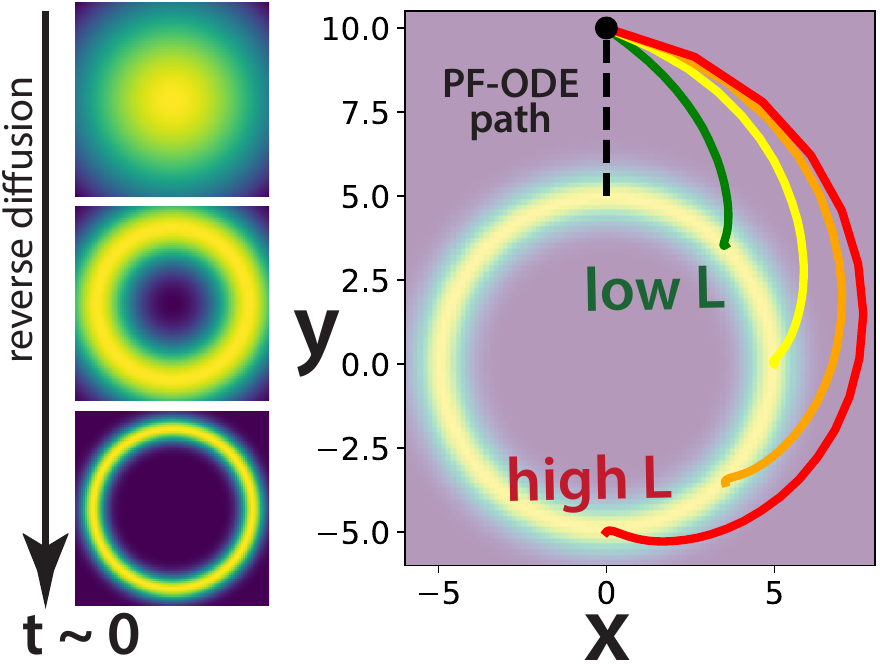}
  \end{center}
  \caption{Angular momentum conservation in reverse diffusion. Left: depiction of reverse diffusion for a rotationally-symmetric data distribution. Right: most likely transition paths given a fixed starting point (black dot) and different angular momentum values (green is lowest, red is highest). Dashed black line is the PF-ODE trajectory.}
\end{wrapfigure}

\section{Conclusion}

The preceding investigation is a proof of principle that Noether's theorem can be usefully applied to study stochastic dynamics, and that the conserved quantities we obtain by doing so (e.g., energy, momentum, angular momentum) recognizably modulate the most likely transitions of systems. As in physics, a system's symmetries constrain its (most likely) behavior. 

Several points suggest interesting directions for future work. First, it appears that Noether's theorem can be used to derive conserved quantities associated with neural networks that are not invariant, but equivariant \citep{Cohen2016-OG}, to at least some transformations; our Lagrangian is rotation-invariant as long as $\vec{f}$ is rotation-equivariant, for example. Second, since the appropriately-formulated converse direction of Noether's theorem is true, it may be possible to learn symmetries from empirically-identified conserved quantities. Finally, given the variety of geometric structures and symmetries relevant to modern AI \citep{Sanborn2024-review}, there are many more conserved quantities to explore beyond energy and momentum.

\clearpage

%%%%%%%%%%%%%%

\bibliography{symbib24}

\end{document}